\documentclass{article}
\usepackage[T1, OT1]{fontenc}
\DeclareTextSymbolDefault{\dh}{T1}

\usepackage{ijcai24}

\usepackage{times}
\usepackage{soul}
\usepackage{url}
\usepackage{xcolor}
\usepackage[colorlinks=true]{hyperref}
\usepackage[utf8]{inputenc}
\usepackage[small]{caption}
\usepackage{graphicx}
\usepackage{amsmath}
\usepackage{amsthm}
\usepackage{comment}
\usepackage{booktabs}
\usepackage{algorithm}
\usepackage{algorithmic}
\usepackage[switch]{lineno}
\usepackage{pifont}

\def \ie {\emph{i.e.}}
\def \eg {\emph{e.g.}}
\def \etal {\emph{et al.}}

\newcommand{\cmark}{\ding{51}}%

\newcommand{\tit}[1]{\smallbreak\noindent\textbf{#1.}}
\newcommand{\tinytit}[1]{\noindent\textbf{#1.}}

\title{Trends, Applications, and Challenges in Human Attention Modelling}

\author{
Giuseppe Cartella$^1$\and
Marcella Cornia$^1$\and
Vittorio Cuculo$^1$\and\\
Alessandro D'Amelio$^2$\and
Dario Zanca${^3}$\and
Giuseppe Boccignone$^2$\And
Rita Cucchiara$^{1}$\\
\affiliations
$^1$University of Modena and Reggio Emilia, Italy\\
$^2$University of Milan, Italy\\
$^3$Friedrich-Alexander-Universitat Erlangen-Nurnberg, Germany\\
\emails
\{giuseppe.cartella, marcella.cornia, vittorio.cuculo\}@unimore.it,
alessandro.damelio@unimi.it,
dario.zanca@fau.de,
giuseppe.boccignone@unimi.it, rita.cucchiara@unimore.it}
\begin{document}

\maketitle

\begin{abstract}
    Human attention modelling has proven, in recent years, to be particularly useful not only for understanding the cognitive processes underlying visual exploration, but also for providing support to artificial intelligence models that aim to solve problems in various domains, including image and video processing, vision-and-language applications, and language modelling. This survey offers a reasoned overview of recent efforts to integrate human attention mechanisms into contemporary deep learning models and discusses future research directions and challenges. 
    For a comprehensive overview on the ongoing research refer to our dedicated repository available at \texttt{\small{\url{https://github.com/aimagelab/awesome-human-visual-attention}}}.
\end{abstract}

\section{Introduction}
\label{sec:intro}
Gaze is a window to the observer's thoughts, intentions, and emotions. It can be used to track attention, identify interests, and even forecast future actions. 
Over the past few decades, extensive research has been conducted, resulting in the proposal of several computational models aimed at predicting attention allocation towards visual and, recently, multimodal stimuli. 
Originally rooted in psychology and neuroscience, these models have garnered increasing interest in the computer vision and pattern recognition community since the publication of the Itti~\etal~\shortcite{itti1998model} model, which provided a viable computational approach to address the problem, paving the way to potential applications in various fields ranging from image and video analysis, robotics, natural language processing, and autonomous driving.

Attention, indeed, plays an essential role in human cognition and survival. Due to the overwhelming amount of incoming sensory information at any given time, evolution has moulded the ability of selective attention. The totality of inputs cannot be fully processed: thus, the brain struggles to select information, allowing for the most effective action to survive in a changing environment. With appropriate differences, the same mechanism is of crucial interest for artificial agents grappling with the complexity of the real world. In this regard, it has become evident that computational models that mirror biological processes can enhance agents' ability to maximise their information intake over time with respect to an internal or external goal. 

In addition, as noted in~\cite{zhang2020human}, the capacity of an AI agent to perceive and interpret human gaze contributes to an efficient and seamless human-machine interaction. 
By deciphering the nuances of gaze patterns, an AI agent gains the ability to engage with users on a deeper level, anticipating their needs and preferences, reacting intuitively to user inputs and improving the overall user experience.

The chief concern of this survey is to offer a reasoned overview of recent efforts to integrate human attention mechanisms into contemporary deep learning models in order to tackle traditional and emerging challenges related to images, text, or multimodal data. Specifically, we explore the integration of gaze data, such as scanpaths or saliency maps, as a strategy to improve performance in various domains, including image and video processing, vision-and-language applications, language modelling, and more specific settings like robotics, autonomous driving, and medicine. 
For instance, in image and video processing, such integration leads to better object recognition, scene understanding, and overall visual comprehension. In the field of vision-and-language applications (\eg~automatic captioning or visual question answering), incorporating gaze data enhances the contextual awareness of the models, allowing for a more refined interpretation of the relationship between visual and textual information. Additionally, the injection of gaze information into language modelling tasks contributes to more contextually informed predictions, improving the model proficiency in reading comprehension and language understanding.
To the best of our knowledge, the perspective of the interplay between deep learning-based applications and visual attention is lacking in most recent reviews of the field. 

This survey is structured as follows. To set out our discussion, Section~\ref{sec:background} provides a brief introduction to the main approaches in human attention modelling, covering both the generation of visual exploration trajectories, also known as scanpath prediction, and the extraction of salient regions in visual stimuli, referred to as saliency prediction. Section~\ref{sec:applications}, the core of this survey, presents a taxonomy of recent studies in application-based solutions that exploit human attention. This is done by considering the type of application, distinguishing between image and video processing, vision and language, and language modelling. Additionally, a focus on applications in specific domains that present extensive literature is provided, such as robotics, autonomous driving, and medicine. Eventually, the open challenges and future directions are discussed in Section~\ref{sec:challenges}.

\section{Human Attention Modelling}
\label{sec:background}
Building upon seminal work on feature integration theory, most computational models of human visual attention assume the central importance of the \textit{saliency map}, \ie~a spatial representation indicating the prominence of information at each location in the visual field. Hence, a \textit{scanpath} of visual attention (\ie~sequence of eye fixation points) is generated by the \textit{winner-take-all} algorithm: the initial attention location is selected as the one with the highest saliency. Subsequent fixations are generated by inhibiting this chosen location and then shifting attention to the next highest saliency location. 
In what follows, we summarise major works in saliency and scanpath prediciton, including models and evaluation metrics.

\subsection{Saliency Prediction}
Based on the assumption of a centralised saliency map, most of the literature focuses on the problem of saliency prediction. Early models mainly exploited image processing techniques based on the identification of features correlated with human attention, often taking inspiration from neuroscience. However, in recent years, large datasets have allowed to learn models directly from human eye-tracking data.

\tit{Models} A seminal implementation of a saliency prediction model was provided by Itti~\etal~\shortcite{itti1998model}. Attention is described as a combination of bottom-up processes, based on conspicuity maps associated with basic features of the visual stimuli like colour, intensity, and edges. These maps are then combined into a single saliency map. 
However, the research by Judd~\etal~\shortcite{judd2009learning} revealed that the sole reliance on bottom-up computation does not accurately align to actual human attention allocation. 
Building upon this intuition, an SVM classifier was trained also on higher-level semantic features contributing a significant performance boost. Jiang~\etal~\shortcite{jiang2015salicon} presented SALICON, a dataset collected by using mouse clicks as a proxy for human visual attention. The introduction of SALICON has represented a significant development in data-driven approaches, leading to the emergence of several deep learning-based architectures. Among them, Cornia~\etal~\shortcite{cornia2018predicting} proposed a convolutional long short-term memory that focuses on the most salient regions of the input image to iteratively refine the predicted saliency map. Lou~\etal~\shortcite{lou2022transalnet} advanced the field by exploring, for the first time, the integration of Transformer-based models within a saliency prediction framework. In particular, a Transformer encoder is built on top of the features extracted by a convolutional network to capture long-range contextual information.

From a different perspective, based on the temporally evolving patterns characterising human attention, Aydemir~\etal~\shortcite{aydemir2023tempsal} defined a model able to perform time-specific saliency prediction. 
Beyond temporal evolution, visual attention exhibits high inter-subject variability, and region prioritisation also depends on the observer's preferences. A related trend~\cite{chen2023learning} has tackled this challenge, laying the foundations for the design of custom applications based on users' interests.

\tit{Evaluation} Saliency evaluation poses inherent challenges, as the comparison between maps is susceptible to numerous factors such as normalisation, smoothing, or centre bias. An in-depth analysis of the principal metrics was carried out in~\cite{bylinskii2019different}, where AUC (Area Under the Curve) and NSS (Normalized Scanpath Saliency) have been established as the most robust. The former provides a comprehensive measure of overall performance, while the latter focuses on the spatial distribution and intensity of predicted saliency relative to human gaze data. 
The most widely recognised benchmarks for saliency prediction were collected for free-viewing conditions on static images, namely MIT1003~\cite{judd2009learning} and SALICON~\cite{jiang2015salicon}\footnote{\texttt{\scriptsize{\url{http://salicon.net/}}}}. For a comprehensive and updated list of datasets collected under different input and task modalities, the reader can refer to the MIT/Tübingen Saliency Benchmark\footnote{\texttt{\scriptsize{\url{https://saliency.tuebingen.ai/}}}}. 

\subsection{Scanpath Prediction}
Unfortunately, by collapsing attention down to saliency maps, the temporal component of the attentional deployment is lost. Two scanpaths that visit analogous locations in an image but in a completely different order generate identical saliency maps. Yet, the exploration order of a visual stimulus is of particular importance since it better reflects the conspicuity of the image elements with respect to the observer, as well as his internal strategy.
Consequently, many studies have been dedicated to modelling scanpaths, in a more sophisticated manner than the previously described winner-take-all algorithm. 

\tit{Models}
The literature on scanpath prediction exhibits a trend similar to saliency prediction, shifting from model-based to deep-learning approaches.
For example, Le Meur~\etal~\shortcite{lemeur_liu} showed that combining eye-movement generation with motor biases, such as saccade amplitudes and orientations, notably improves the quality of the generated scanpath. Differently, Zanca~\etal~\shortcite{zanca2019gravitational} conceived a visual attention scanpath as the motion of a unitary mass within gravitational fields generated by salient visual features. Recent approaches have exploited the semantics of features learned by deep neural networks to improve modelling.
K{\"u}mmerer~\etal~\shortcite{kummerer2022deepgaze} introduced DeepGaze III, which entangles a spatial priority network that outputs a priority map, with a scanpath network conditioned on the history of previous fixations. 

While most of the research has concentrated on developing  models in free-viewing conditions, the study of goal-directed attention (\eg~visual search tasks) is still under-explored. Mondal~\etal~\shortcite{mondal2023gazeformer} have opened up new avenues in this direction. Their model encodes the target by adopting a natural language model, thus leveraging semantic similarities in scanpath prediction. The training is devised to solve a zero-shot task to recognise never-before-searched objects.

\begin{figure*}[t!]
    \centering
    \includegraphics[width=0.99\linewidth]{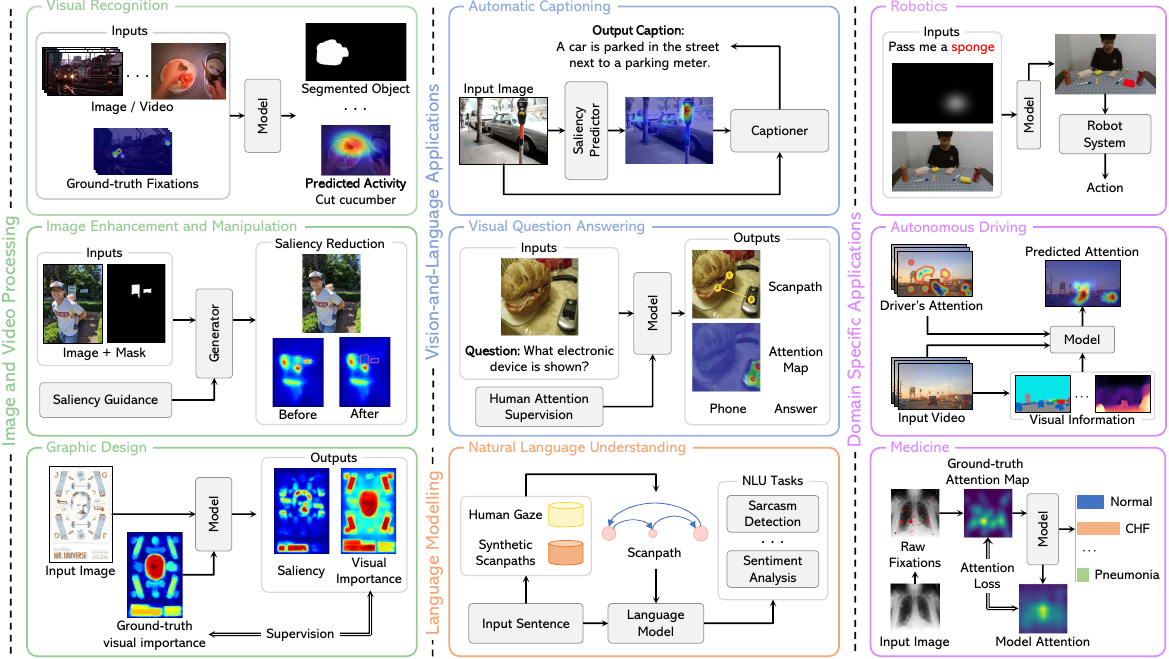}
    \vspace{-0.1cm}
    \caption{An overview of sample architectures 
    integrating human visual attention with different input and output modalities. Human visual attention 
    has been employed to solve tasks in diverse domains spanning from image and video processing, automatic captioning, visual question answering, and language understanding, as well as robotics, autonomous driving, and medicine.}
    \label{fig:main-figure}
    \vspace{-0.25cm}
\end{figure*}

\tit{Evaluation} Different from the case of saliency prediction, when assessing the similarity between visual attention scanpaths, accounting for the temporal order of fixation and the high inter-subject variability is crucial. The string-edit distance proved to be a robust metric for comparing scanpaths after they had been appropriately transformed into strings. For a thorough exploration of scanpath prediction metrics, we direct the reader's attention to~\cite{kummerer2023predicting}.
Some visual attention benchmarks provide information about individual fixations, allowing the evaluation of scanpath models. Among those, the aforementioned MIT1003 and COCO-FreeView\footnote{\texttt{\scriptsize{\url{https://sites.google.com/view/cocosearch/coco-freeview}}}} are two examples of datasets containing eye fixations collected under the free-viewing task. 

\section{Integrating Human Attention in AI Models}
\label{sec:applications}

Visual attention modelling has recently evolved into a valuable tool supporting state-of-the-art AI-based approaches across various domains, extending from computer vision to natural language processing. Recent literature reflects a non-negligible increase in leveraging human attention guidance, obtained from eye-tracking data or computational models, to address a wide range of both established and emerging challenges (see Figure~\ref{fig:main-figure}). An intriguing development is the integration of human attention principles beyond visual stimuli. This expansion encompasses tasks ranging from vision-only to those involving text, where attention modelling incorporates text gaze patterns.
The subsequent section explores and summarises approaches benefiting from human attention adoption, offering insights into its broad applicability. Table~\ref{tab:summary} provides a concise overview.

\subsection{Image and Video Processing} 
\tinytit{Visual Recognition} Human attention modelling has been successfully adopted in a variety of computer vision-related tasks such as object detection~\cite{fan2019shifting} and segmentation~\cite{wang2019learning} or activity recognition~\cite{min2021integrating}. The recorded human gaze data assists AI models in detecting salient objects in videos, segmenting videos without supervision and recognising activities in egocentric video sequences. In~\cite{qiao2023joint}, human visual attention is employed in a multi-task learning framework for audio–visual saliency prediction and sound source localisation, proving that attention is a valuable guide to pinpoint the origin of sounds in videos. Other notable examples relate to the concept of foveated processing, \ie~prioritising computations around a focal point as in the human visual system. Foveated processing has been recently introduced as a novel structural component of modern deep learning architectures, delivering significant inference speed-ups compared to classic convolutional networks and competitive results on different image and video recognition tasks~\cite{tiezzi2022foveated}.

\tit{Graphic Design} One compelling application of human attention modelling is related to the automatic evaluation of design choices of websites, data visualisations and infographics, mobile UIs, or advertisements and posters~\cite{bylinskii2017learning,leiva2020understanding}.
Notably, visual attention allocation in graphic designs can be interpreted as a proxy for the perceived relative importance of design elements. Being able to effectively and automatically perform such assessment paves the way to a spectrum of downstream applications, including automatic image retargeting and thumbnailing~\cite{bylinskii2017learning} or interactive design~\cite{fosco2020predicting}.  
Recent approaches, while keeping the neural architectural structure of saliency models, are trained to predict \textit{importance maps}. Adopted datasets and ground-truth importance maps are collected either via BubbleView mouse clicks~\cite{bylinskii2017learning} or by manual labelling of important regions~\cite{fosco2020predicting}.

\tit{Image Enhancement and Manipulation} Visual attention allocation models, particularly saliency models, have shown effectiveness in various tasks aimed at manipulating images to enhance the attractiveness of specific objects therein. Such tasks, sometimes referred to as attention retargeting or re-attentionizing, aim at automatically enhancing (or weakening) some portions of the original stimulus, thus reducing visual distraction and highlighting the desired object. In~\cite{gatys2017guiding}, this is achieved by training a convolutional neural network to transform images so as to satisfy a given fixation distribution eventually predicted by a saliency model.
Target fixation distribution is obtained by either increasing or decreasing the saliency of particular objects or globally scaling the saliency map. Similarly, in~\cite{jiang2021saliency}, the authors proposed a saliency-guided image translation model, named SalG-GAN, designed for image-to-image translation based on user-specified saliency maps. SalG-GAN employs a disentangled representation framework to encourage diverse translations for the same saliency and introduces a saliency-based attention module to enhance the key components. 

More recently, Aberman~\etal~\shortcite{aberman2022deep} proposed a method to reduce distractions in images by utilising a saliency model. By inputting an image and a mask specifying the editing region, the approach employs backpropagation to parameterise a differentiable editing operator, diminishing saliency in the edited area. Various operators are defined, encompassing a recolouring operator for seamless integration of distractors, a warping operator to gradually eliminate objects, and a generative adversarial network operator to substitute regions with less salient alternatives using a semantic prior. Similarly, in~\cite{miangoleh2023realistic}, a realistic saliency-guided image enhancement method was proposed to maintain high manipulation realism while attenuating distractors and amplifying objects of interest. To this end, a realism loss for saliency-guided image enhancement was designed for training the system to optimise the saliency of a region while penalising for deviations from realism.

\tit{Image Quality Assessment (IQA)} IQA involves using computational models to measure image quality in alignment with subjective human evaluations. As significant distortions can impact visual attention allocation on a given stimulus, saliency and scanpath models have been adopted to increase the effectiveness of perceptual IQA methods.

One recent example is SGDNet~\cite{yang2019sgdnet}, an end-to-end deep neural network for no-reference IQA. It optimises visual saliency and image quality prediction employing a multi-task learning framework. The authors highlight how this approach can leverage the correlation between saliency information and image quality, utilising saliency maps and quality scores during training. Handling saliency prediction as a sub-task enhances feature fusion, thus producing more perceptually consistent outcomes. Zhu~\etal~\shortcite{zhu2021saliency} introduced TranSLA, a saliency-guided Transformer network for no-reference IQA. TranSLA jointly learns to predict mean opinion scores on a given image and the corresponding saliency map. The latter is used as the query value of the multi-head self-attention mechanism of a Transformer architecture, while the key-value pair is provided by the input image feature representation from a ResNet-based convolutional network. Eventually, this forces the model to focus on the most salient regions, employing a pre-trained saliency model to provide ground-truth saliency maps.
Recent studies have demonstrated that viewing behaviours (\ie~scanpaths) are useful for perceptual IQA on $360^{\circ}$ images and videos, where eventual distortions are localised in space and time. In \cite{sui2023scandmm}, a scanpath prediction model on $360^{\circ}$ images is employed to sample a sequence of rectilinear projections of viewports (\ie~frames) along different generated scanpaths. Quality scores are then computed on each frame. Video-level quality scores are obtained by temporally pooling the frame-level quality scores. The proposed approach shows better performance compared to naively applying IQA on equirectangular projections.

\begin{table}[t]
\centering
\setlength{\tabcolsep}{.25em}
\resizebox{\linewidth}{!}{
\begin{tabular}{lc cccc c cc c l}
\toprule
& & \multicolumn{4}{c}{\textbf{Input}} & & \multicolumn{2}{c}{\textbf{Human Attention}} \\
\cmidrule{3-6} \cmidrule{8-9}
\textbf{Models} & & I & V & A & T & & \textbf{Type} & \textbf{Real} & & \textbf{Task} \\
\midrule
\cite{wang2019learning} & & & \cmark & & & & Fixation points & \cmark & & Visual Recognition \\
\cite{min2021integrating} & & & \cmark & & & & Fixation points & \cmark & & Visual Recognition \\
\cite{qiao2023joint} & & & \cmark & \cmark & & & Saliency maps & \cmark & & Visual Recognition \\
\midrule
\cite{bylinskii2017learning} & & \cmark & & & & & Saliency maps & \cmark & & Graphic Design \\
\cite{fosco2020predicting} & & \cmark & & & & & Saliency maps & \cmark & & Graphic Design \\
\cite{leiva2020understanding} & & \cmark & & & & & Saliency maps & \cmark & & Graphic Design \\
\midrule
\cite{jiang2021saliency} & & \cmark & & & & & Saliency maps & \cmark & & Image Manipulation \\
\cite{aberman2022deep} & & \cmark & & & & & Saliency maps & & & Image Manipulation \\
\cite{miangoleh2023realistic} & & \cmark & & & & & Saliency maps & & & Image Manipulation \\
\midrule
\cite{yang2019sgdnet} & & \cmark & & & & & Saliency maps & & & IQA \\
\cite{zhu2021saliency} & & \cmark & & & & & Saliency maps & & & IQA \\
\cite{sui2023scandmm} & & \cmark & & & & & Scanpaths & & & IQA \\
\midrule
\cite{sugano2016seeing} & & \cmark & & & \cmark & & Fixation points & \cmark & & Captioning \\
\cite{yu2017supervising} & & & \cmark & & \cmark & & Saliency maps & & & Captioning \\
\cite{cornia2018paying} & & \cmark & & & \cmark & & Saliency maps & & & Captioning \\
\midrule
\cite{qiao2018exploring} & & \cmark & & & \cmark & & Saliency maps & & & VQA \\
\cite{sood2023multimodal} & & \cmark & & & \cmark & & Saliency maps & & & VQA \\
\cite{ilaslan2023gazevqa} & & & \cmark & & \cmark & & Saliency maps & \cmark & & VQA \\
\midrule
\cite{berzak2017predicting} & & & & & \cmark & & Fixation points & \cmark & & Reading Comprehension \\
\cite{reich2022inferring} & & & & & \cmark & & Scanpaths & \cmark & & Reading Comprehension \\
\cite{skerath2023native} & & & & & \cmark & & Fixation points & \cmark & & Reading Comprehension \\
\midrule
\cite{sood2020improving} & & & & & \cmark & & Saliency maps & & & Language Understanding \\
\cite{khurana2023synthesizing} & & & & & \cmark & & Scanpaths &  & & Language Understanding \\
\cite{deng2023pre} & & & & & \cmark & & Scanpaths & & & Language Understanding \\
\bottomrule        
\end{tabular} 
}
\vspace{-0.1cm}
\caption{A summary of representative works incorporating human attention across various tasks. We outline the adopted modalities (\textit{I}: Image, \textit{V}: Video, \textit{A}: Audio, \textit{T}: Text), how human attention is processed (via \textit{saliency maps}, full \textit{scanpaths}, or raw \textit{fixation points}), and whether the model employs \emph{real} human attention data from eye-trackers or predictions from computational models.}
\label{tab:summary}
\vspace{-0.3cm}
\end{table}

\subsection{Vision-and-Language Applications}

\tinytit{Automatic Captioning} 
Within the domain of tasks related to vision and language, automatic captioning has gained significant attention as an effective way of generating natural language descriptions for both images and videos. Captioning models are designed not only to identify all pertinent objects and entities within a scene but also to generate coherent and grammatically correct sentences describing them. While the advancements in this field are promising, some works have explored the possibility of integrating human gaze information to enhance the generation process of textual descriptions.

The first attempt in this direction was presented by Sugano and Bulling~\shortcite{sugano2016seeing}, in which fixation points are used as an additional input to guide the caption generation of a recurrent-based image captioning model. In particular, they integrated gaze information, represented in the form of fixation histograms, at each time step of the recurrent neural network where fixated and non-fixated regions are combined to produce the output sentence. This approach is limited by the need to have both eye fixation and caption annotations for the same visual input. As a solution, Cornia~\etal~\shortcite{cornia2018paying} shifted to the use of saliency maps predicted by a pre-trained visual saliency model instead of employing raw fixation points, thus being able to apply the proposed approach to any input image and potentially extend training to any publicly available captioning dataset. Moreover, given that humans tend to pay attention not only to the salient regions but also to the surrounding context when describing an image, they also proposed to let the model learn how to weigh the contribution of both salient and contextual visual features during the generation of the caption. 

On a similar line, other works~\cite{yu2017supervising} employed predicted saliency maps in conjunction with additive attention mechanisms to effectively capture all semantic cues typically described in a textual sentence. While previous methods take images as input, Yu~\etal~\shortcite{yu2017supervising} proposed to employ saliency information directly predicted inside the model for describing video sequences. He~\etal~\shortcite{he2019human}, instead, extracted fixated locations from predicted saliency maps via the winner-take-all approach and used them to obtain a set of visual features, each corresponding to the spatial image region centred on a fixated location, that is given as input to the language model. Additionally, they introduced a novel dataset for the task composed of 4,000 images with corresponding eye movements and verbal descriptions recorded synchronously over images. Similar annotations were employed in the model proposed by Takmaz~\etal~\shortcite{takmaz2020generating}, in which a saliency map is extracted for each word of the caption, taking advantage of the alignment between eye fixations and recorded audio traces with image descriptions. This sequence of saliency maps is used as an additional input for the captioning model, using the saliency map corresponding to the word to be predicted at each time step of the generation.

\tit{Visual Question Answering (VQA)}
In recent advancements within vision-and-language research, there has been a growing emphasis on integrating human attention to enhance the performance of VQA models, which aim to effectively answer questions based on a given image or video sequence. In this context, Das~\etal~\shortcite{das2016human} conducted a preliminary investigation and introduced the VQA-HAT dataset, comprising image-question pairs and corresponding attention maps collected through a mouse-based procedure. Their evaluation of attention maps from state-of-the-art VQA models against human attention revealed a discrepancy between machine-generated and human attention, highlighting the need for enhanced attention mechanisms. A subsequent work~\cite{qiao2018exploring} employed the VQA-HAT dataset to address the scarcity of gaze-based data in VQA. Specifically, this study proposed a network capable of generating human-like attention maps, which are employed to enhance attention-based VQA models with human-like attention supervision, resulting in significant improvements. Chen~\etal~\shortcite{chen2020air} continued in this direction by leveraging human attention to promote reasoning behaviour in a VQA model and collected an eye-tracking dataset to enhance the understanding of human attention during the visual reasoning process.

In a different vein, Sood~\etal~\shortcite{sood2021vqa} presented the VQA-MHUG dataset, focusing on multimodal human gaze on both images and questions in VQA. Their analysis revealed a significant correlation between human attention on text and VQA performance across various models, underscoring the importance of text attention in understanding visual scenes. Subsequent research~\cite{sood2023multimodal} adopted this dataset and proposed a VQA model that integrates human-like attention on both image and text during training, showing competitive performance against previous methods.

While all the aforementioned approaches were designed to work on images, Ilaslan~\etal~\shortcite{ilaslan2023gazevqa} shifted towards video question answering and introduced a task-oriented VQA scenario for collaborative tasks. Notably, their contribution involved the creation of a dataset with human gaze data collected for first-person view videos and the introduction of a novel approach for task-specific question answering, emphasising the importance of gaze information in understanding user intent during collaborative tasks.

\subsection{Language Modelling}

\tinytit{Machine Reading Comprehension} Advancements in machine reading comprehension, particularly in predicting native language from gaze patterns, have been a focal point in recent natural language processing research. Berzak~\etal~\shortcite{berzak2017predicting} introduced an innovative methodology for studying cross-linguistic influence by analysing eye movement patterns in second-language reading. The proposed framework demonstrated the predictive power of gaze fixations, revealing a systematic influence of native language properties on reading in a second language. Building on this, Skerath~\etal~\shortcite{skerath2023native} conducted a study to validate the identification of native language from gaze records. By utilising a new corpus with a larger set of native languages, they reinforced the robustness of the approach, showcasing comparable classification performance with reduced data. Reich~\etal~\shortcite{reich2022inferring}, instead, designed the first LSTM-based model capable of inferring reading comprehension, text difficulty, and the reader's native language from scanpaths in reading. Their model, which incorporates lexical, semantic, and linguistic features, treats eye movements as an independent variable, highlighting the potential to infer reader properties comprehensively.

\tit{Natural Language Understanding}
The integration of human attention into natural language understanding tasks has recently emerged as a promising avenue for improving the performance of various models. In this context, the work by Sood~\etal~\shortcite{sood2020improving} stood out for adopting human gaze information into neural attention mechanisms for natural language processing tasks. Faced with data scarcity challenges, the authors proposed a novel hybrid text saliency model, combining a cognitive model of reading behaviour with human gaze supervision. The proposed model was further integrated into an attention layer, demonstrating substantial performance gains in paraphrase generation and sentence compression tasks.  

Khurana~\etal~\shortcite{khurana2023synthesizing} contributed to this field with the introduction of a model capable of generating scanpaths from existing eye-tracking corpora. The generated scanpaths proved beneficial for various tasks, such as sentiment classification, paraphrase detection, and sarcasm detection. Following the same direction, Deng~\etal~\shortcite{deng2023pre} leveraged synthetic gaze data and developed a model that combined scanpath generation with a BERT-based language model, eliminating the need for human gaze data. The proposed approach outperforms the underlying language model in multiple tasks, especially in low-resource settings, showcasing the potential of synthetic eye-gaze data.

\subsection{Domain-Specific Applications}

\tinytit{Robotics}
Human-robot interaction has witnessed a paradigm shift towards the integration of socially intelligent robots into various facets of human life.
When it comes to patients with motor disabilities, assistive robotics makes a fundamental contribution to daily tasks, making the surrounding environment more inclusive. In this context, the human gaze represents a fundamental nonverbal communication channel used to convey our intentions, goals, or preferences, while offering a natural and non-invasive human-machine interface. Robot assistance can be enabled by individuals directing their gaze towards a specific target they wish to grasp or manipulate. As emphasised in~\cite{qian2023gvgnet}, such an approach allows humans to convey their object-referring intentions without ambiguity.

In the specific domain of collaborative robotics, 
effective interaction and mutual understanding between humans and robots is essential for the successful accomplishment of a given task.
In early teleoperation applications, robots were controlled through joystick input signals. However, this control mode poses inherent complexities and demands substantial human expertise. Furthermore, joystick inputs constrain the capacity to anticipate user intentions by conveying only incremental instructions, thus limiting the system's awareness of the ultimate goal.
In response to these limitations, researchers have embarked on the development of shared control systems able to accommodate eye-gaze input modalities, thereby diverging from the exclusive reliance on joystick signals~\cite{aronson2022gaze}.

Another fundamental field in the realm of robotics is related to the process of a robot acquiring knowledge through the imitation of an expert, also known as learning from demonstrations. 
Saran~\etal~\shortcite{saran2019understanding} characterised the gaze patterns of human instructors during task demonstrations and observed a tendency among users to disregard task-irrelevant objects.
Their work proved that integrating gaze information with inverse reinforcement learning enhances the robot's ability to learn reward structures.
Huang~\etal~\shortcite{huang2019nonverbal} attempted to model robot gaze as a feedback signal intended to communicate with humans and enhance the efficacy of human instructions and the overall quality of the learning process.

\tit{Autonomous Driving}
Developing intelligent agents able to perceive the environment and autonomously perform actions is a primary objective of AI research. In this context, the design of autonomous vehicles is one of the most striking examples. The human visual attention system is unique in this respect. Human beings have the innate ability to rapidly discern the most relevant stimuli, filter out unnecessary information and, while driving, promptly recognise and preempt potential risks.
Transferring these inherent human abilities onto autonomous vehicles constitutes an intricate challenge that has attracted the interest of the research community.

Palazzi~\etal~\shortcite{palazzi2018predicting} pioneered a significant endeavour in the field by introducing one of the earliest attempts, in the deep learning era, to model driver's attention.
With this aim, they presented the first publicly available dataset, named \texttt{DR(eye)VE}, which features in-car recordings under different traffic and weather conditions, as well as variegate landscapes. Accurately modelling the behaviour of autonomous agents in critical circumstances, such as collisions and accidents, represents a major concern for the safety and security of passengers. Nonetheless, collisions are quite rare, therefore the in-car protocol delineated by Palazzi~\etal~\shortcite{palazzi2018predicting} makes the modelling of such events nearly impossible due to the scarcity of data. 

As a consequence, Xia~\etal~\shortcite{xia2018predicting} and Baee~\etal~\shortcite{baee2021medirl} presented an in-lab data gathering protocol to simulate accident-prone scenarios, involving collisions, near-collisions, and braking events. These efforts resulted in the creation of the BDD-A and EyeCar datasets respectively. Noteworthy, the model proposed in~\cite{xia2018predicting} demonstrated generalisation to real-world scenarios, even when trained on in-lab data only.
Baee~\etal~\shortcite{baee2021medirl} presented a model based on an inverse reinforcement learning schema that predicts an attention policy treating each fixation point as a potential source of reward. An additional distinguishing feature of the human visual system is the peripheral vision. It denotes the capacity to focus our attention on one object while concurrently attending to another. Pal~\etal~\shortcite{pal2020looking}
took into consideration this particular capability and introduced a gaze detection model augmented by the semantics of the scene, comprising the most relevant objects to the driving task, as well as distances and speed.

\tit{Medicine}
In recent years, the advancement of sophisticated deep-learning techniques has notably contributed to the field of medical image analysis, consequently leading to the success of computer-aided diagnosis techniques. Conventional methodologies usually employ convolutional or Transformer-based architectures to learn the textural pattern of diseases and the spatial relationships among image patches.

Medical images exhibit intricate anatomical structures. This poses challenges for a deep network, uniquely based on its inner representations, to precisely capture diagnostically relevant regions. Moreover, conventional models lack the required domain knowledge to correctly diagnose a disease in a more interpretable manner, while avoiding spurious correlations~\cite{ma2023eye}. In the medical field, practitioners often rely on visual information, and the outcome of a diagnosis is closely linked to the observation and interpretation of a medical image. Here, human gaze emerges as a natural way to capture visual attention during the diagnosis process, and motivates recent research towards the implementation of human attention-augmented architectures for the healthcare domain. Advancements in this direction have been facilitated by the introduction of medical datasets that incorporate eye-gaze data obtained from human experts during the diagnostic process. Karargyris~\etal~\shortcite{karargyris2021creation} released a dataset of medical images equipped with raw gaze annotations provided by radiologists, complemented by bounding boxes marking the discriminative areas.
Noteworthy, the authors highlighted the significance of acquiring fine-grained data extending beyond basic labels representing disease classes. 
Indeed, more detailed annotations lead to better localisation capabilities,
thereby enhancing interpretability, an essential requirement for high-risk applications.

Building upon this contribution, several models trained on gaze medical data have been developed. From an architectural point of view, two principal research directions can be identified. The first entails the conversion of raw gaze data into a visual attention map, employing it as a source of supervision to enhance the alignment of model attention with human attention.
Bhattacharya~\etal~\shortcite{bhattacharya2022radiotransformer} proposed a teacher-student framework in conjunction with a visual attention loss, guiding the student attention towards regions identified by the attention map generated by the teacher, that was trained on the ground-truth visual search patterns.
Concurrently, Wang~\etal~\shortcite{wang2022follow} defined an attention-consistency module, suppressing inter-observer variability. The second line of research, instead, is based on a two-stream architecture. The input image and the related attention map are typically processed separately before being concatenated to predict the final output. 
In particular, Ma~\etal~\shortcite{ma2023eye} proposed a gaze-guided Vision Transformer that receives image patches as input, appropriately masked to exclude irrelevant regions for the diagnosis.

In contrast to the previously outlined research directions, Wang~\etal~\shortcite{wang2024gazegnn}, proposed for the first time to directly embed raw gaze data into a graph neural network for X-ray classification. This architecture avoids the adoption of visual attention maps as commonly defined in more classical models.
\section{Open Challenges and Future Directions}
\label{sec:challenges}
We have so far discussed how comprehending human attention represents a multifaceted endeavour and its observable aspects, such as gaze deployment via eye movements, serve as valuable indicators for improving the performance and explainability of AI-related tasks. However, incorporating this information into AI models still represents a major challenge. 

\tit{Data scarcity}
One possible reason for this is the scarcity and cost of collecting human gaze data. When it comes to training machine learning models that rely on human attention, the shortage of data poses a significant challenge. One possible solution is to try to reduce the cost of eye-tracking hardware and make data collection less dependent on human instructors.
This might be fostered by the increased interest and proliferation of wearable devices, such as smart glasses, virtual and augmented reality headsets, which would allow large amounts of data to be collected in an easy way and, most importantly, in an ecological setting. However, this is likely to raise a new challenge concerning ethical and privacy-aware data collection and sharing.

\tit{Privacy issues}
Currently, there is a lack of research into the privacy issues raised by eye tracking on a broad scale~\cite{gressel2023privacy}. To better understand the privacy risks inherent in naturalistic real-world situations, it is essential to focus on eye-tracking datasets that document extensive gaze data during typical daily activities. A notable example of such datasets is Ego4D~\cite{grauman2022ego4d}. This contains 3,670 hours of daily-life activity videos together with eye gaze data. This approach is unique compared to most research datasets. The analysis involves examining the real-world implications of eye tracking and exploring the dynamics of privacy-sensitive situations that may occur during everyday activities. The focus is on actual situations rather than artificial scenarios. By carefully examining these instances and assessing the level of privacy risks from the user's perspective, it would be possible to gain valuable insights into the immediate concerns related to eye tracking. This can establish a proactive understanding of potential long-term risks, enabling the development of strategies and safeguards to anticipate, mitigate, and prevent privacy-related challenges in the evolving landscape of eye-tracking technologies.

\tit{Synthetic data}
Another way to address both the aforementioned problem and that of data scarcity is to use synthetic eye movements to supplement the existing data.
Researchers have developed models that integrate synthetic scanpath generation with a scanpath-augmented language model, eliminating the need for human gaze data~\cite{khurana2023synthesizing,deng2023pre}.
This approach has shown promising results for natural language processing tasks, while minimal exploration has been undertaken in computer vision.

\tit{Wearable devices}
Back to the spread of head-mounted displays like augmented and virtual reality headsets, by adapting the capabilities of vision-and-language models to stimuli obtained in real-time from the surrounding world, it would be possible to enhance the user experience and improve the applicability and effectiveness of AI agents in everyday scenarios~\cite{yan2023voilaa}.
This integration enables a focus on user preferences, providing precise and relevant responses by understanding intent, context, and individual interests. This approach is particularly beneficial for visually impaired individuals who rely on their gaze to communicate their intent. 

Eventually, other related research areas present challenges that still need to be addressed, such as multimodal gaze following, reliable real-time gaze prediction, and gaze modelling for dynamic environments~\cite{ghosh2023automatic,hou2024multi}. Efforts spent in these realms have the potential to significantly advance the field of human gaze modelling and enable the development of more effective applications.

\section*{Acknowledgments}
This work was partially supported by the PNRR project Italian Strengthening of Esfri RI Resilience (ITSERR) funded by the European Union - NextGenerationEU (CUP: B53C22001770006).
\clearpage
\bibliographystyle{named}
\bibliography{ijcai24}

\end{document}